\documentclass{bmvc2k}

\usepackage{times} 
\usepackage{amsmath} 
\usepackage{amssymb}  

\usepackage{graphicx}
\usepackage{algorithm,algpseudocode}
\usepackage{csquotes}
\usepackage{color}
\usepackage{multirow}
\usepackage{upgreek}
\usepackage{booktabs}

\newcommand{\vek}[1]{\mathbf{#1}}
\newcommand{\veksub}[2]{\mathbf{#1}_\text{#2}}
\newcommand{\matsub}[2]{\mathbf{#1}_\text{#2}}

\newcommand{\proba}[1]{\text{p}(#1)}
\newcommand{\probahat}[1]{\hat{\text{p}}(#1)}
\newcommand{\good}[1]{\textit{#1}}
\newcommand{\bad}[1]{\color{red}{#1}}
\newcommand{\best}[1]{\textbf{#1}}

\title{Ki-Pode: Keypoint-based Implicit Pose Distribution Estimation of Rigid Objects}

\addauthor{Thorbj{\o}rn Mosekj{\ae}r Iversen}{thmi@mmmi.sdu.dk}{1}
\addauthor{Rasmus Laurvig Haugaard}{rlha@mmmi.sdu.dk}{1}
\addauthor{Anders Glent Buch}{anbu@mmmi.sdu.dk}{1}

\addinstitution{
 SDU Robotics\\
 The Maersk Mc-Kinney Moller Institute\\
 University of Southern Denmark, DK\\
}

\runninghead{Iversen, Haugaard, Buch}{Ki-Pode: Keypoint-based Impl. Pose Dist. Est.}

\begin{document}

\maketitle


\begin{abstract}
The estimation of 6D poses of rigid objects is a fundamental problem in computer vision. Traditionally pose estimation is concerned with the determination of a single best estimate. However, a single estimate is unable to express visual ambiguity, which in many cases is unavoidable due to object symmetries or occlusion of identifying features. Inability to account for ambiguities in pose can lead to failure in subsequent methods, which is unacceptable when the cost of failure is high. Estimates of full pose distributions are, contrary to single estimates, well suited for expressing uncertainty on pose. Motivated by this, we propose a novel pose distribution estimation method. An implicit formulation of the probability distribution over object pose is derived from an intermediary representation of an object as a set of keypoints. This ensures that the pose distribution estimates have a high level of interpretability. Furthermore, our method is based on conservative approximations, which leads to reliable estimates. The method has been evaluated on the task of rotation distribution estimation on the YCB-V and T-LESS datasets and performs reliably on all objects.
\end{abstract}
\section{Introduction}
\label{sec:introduction}
Many robotics applications which involve the manipulation of rigid objects require accurate knowledge of object poses to succeed. The topic of pose estimation has, thus, received considerable attention from the computer vision community.  While the literature on pose estimation covers many different sensor modalities, pose estimation from an RGB image remains highly relevant due to the low cost of industry-grade RGB cameras.

The majority of the literature on pose estimation is concerned with the estimation of a single best estimate of object pose. However, a single estimate is inherently ill-suited to express visual ambiguity which is unavoidable in many practical applications due to occlusions, poor image and lighting conditions, or object symmetries. In recent years there has, therefore, been an increasing interest in the estimation of probability distributions over object pose.

Distribution estimates are especially valuable when used in contexts with a high cost of failure such as robotic grasping. In such contexts, it is favorable to postpone a task until the object's pose is known with adequate accuracy and precision. Overconfident estimates of uncertainty can, however, still lead to failure, so it is crucial that the estimates are reliable, in the sense that the false positive rate is low. Underconfident estimates simply risk being so uncertain that subsequent tasks are not performed, even when the information necessary for successful completion was available. However, this can be alleviated through sensor fusion with additional views which the distribution estimates are well suited for.

Estimating a full 6D probability distribution over object pose is non-trivial, and most prior work focuses on rotation distribution estimation, modeling the uncertainty either as a mixture of parametric distributions or as a probability mass function over sampled rotations in SO(3). We argue that both methods are somewhat limited in their descriptive ability, restricted by the number of distributions or the number of samples, respectively.

\begin{figure}[bt]
\centering
\includegraphics[width=0.15\columnwidth]{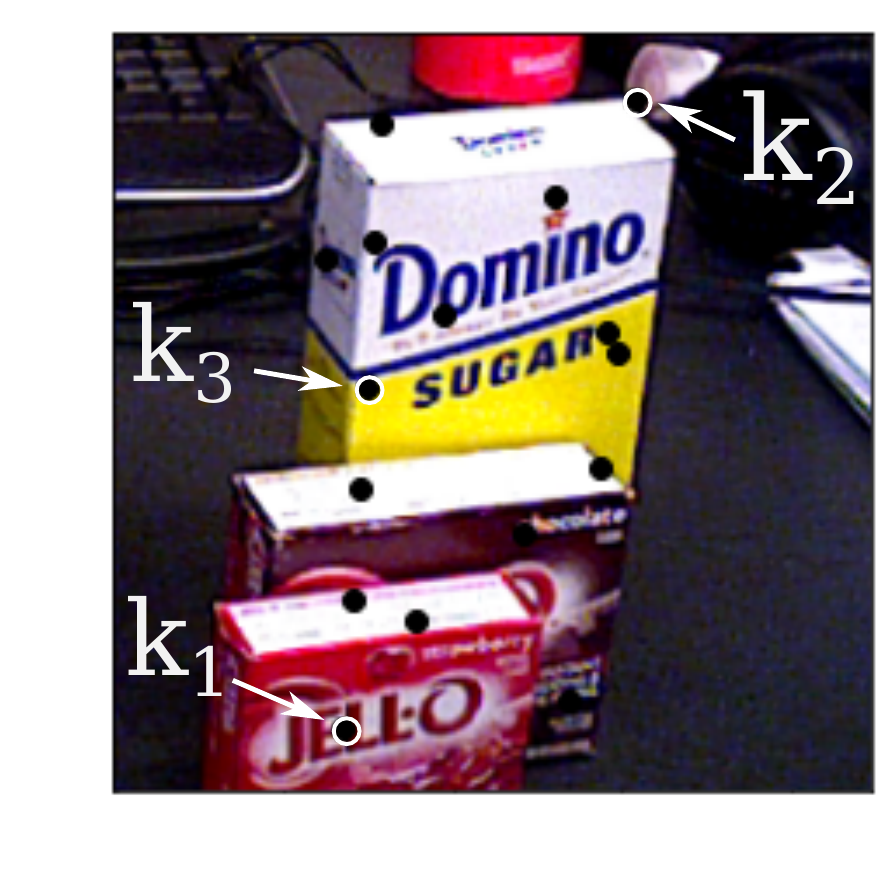}
\includegraphics[width=0.46\columnwidth, trim={0 185 860 0},clip]{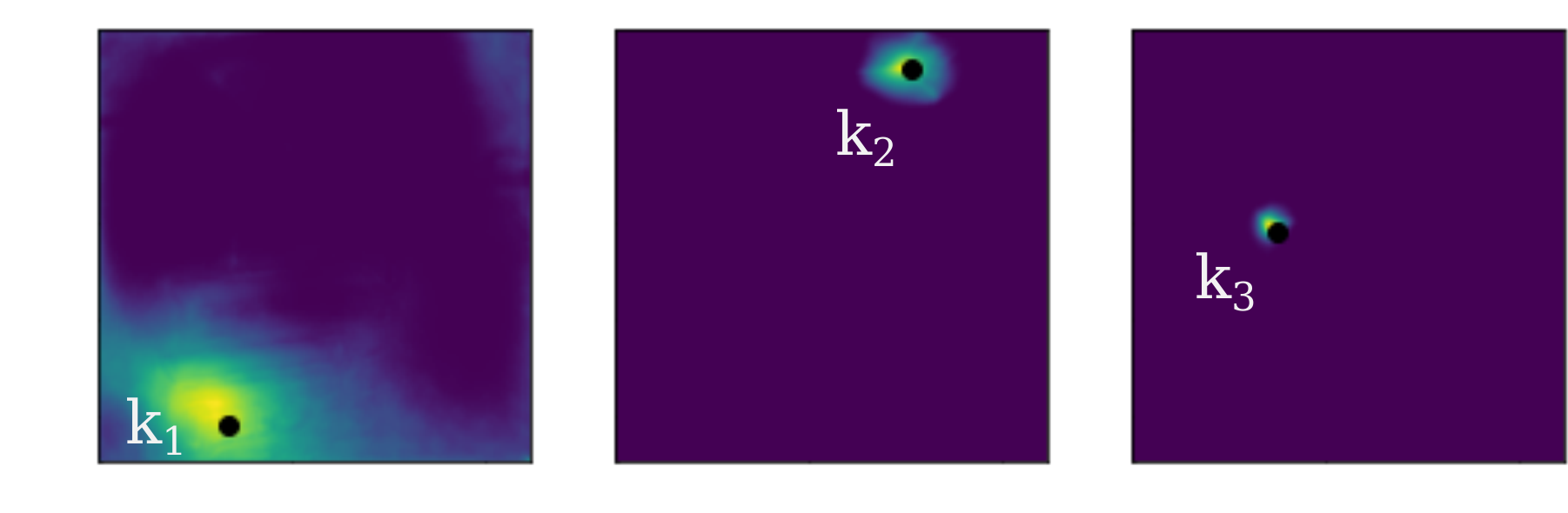}
\includegraphics[width=0.3\columnwidth]{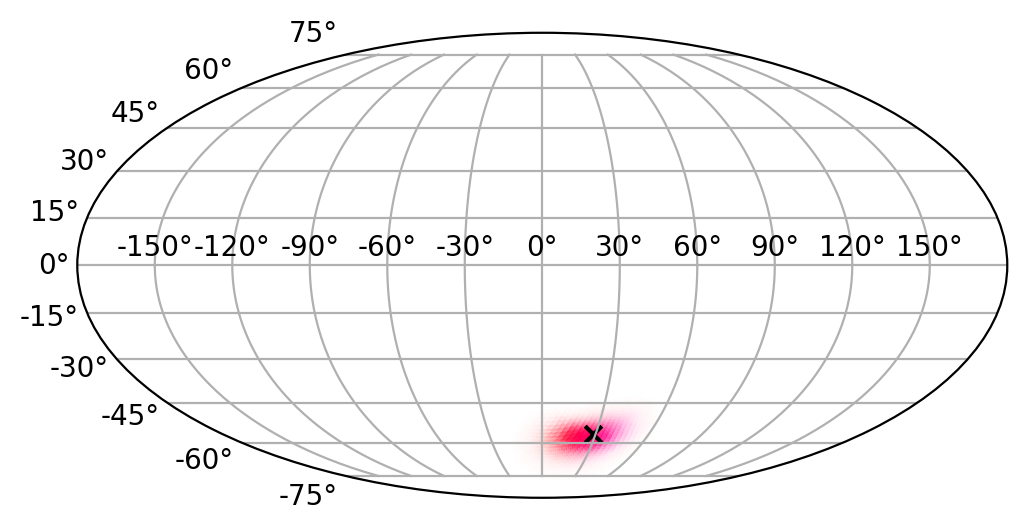}

\includegraphics[width=0.15\columnwidth]{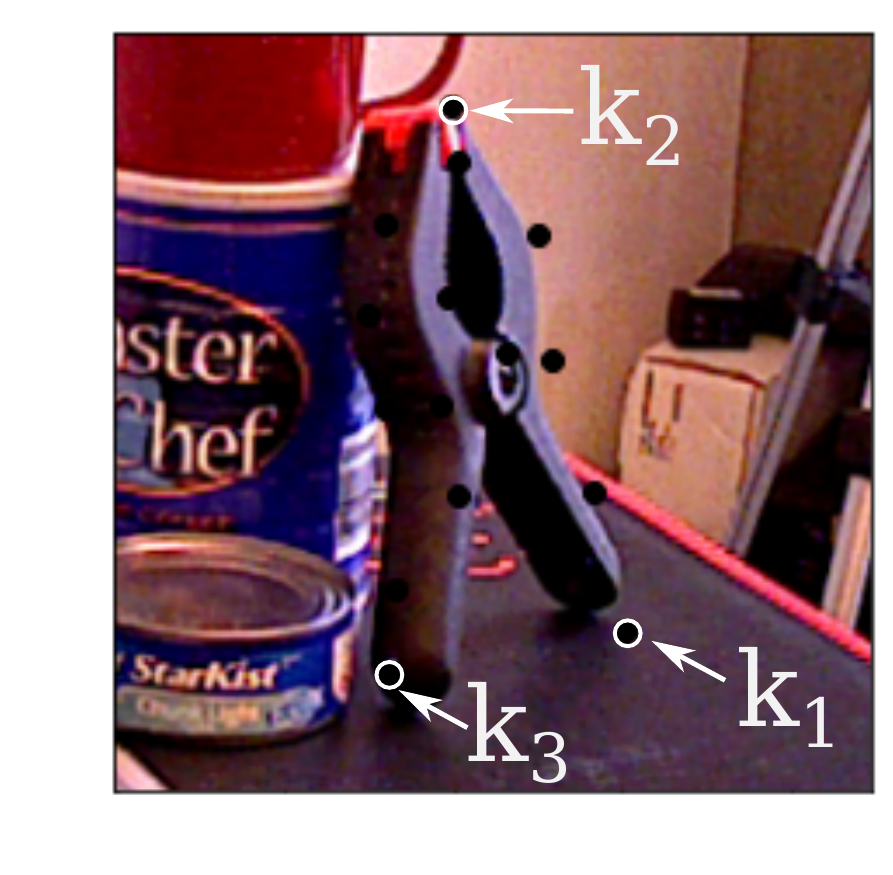}
\includegraphics[width=0.46\columnwidth, trim={0 185 860 0},clip]{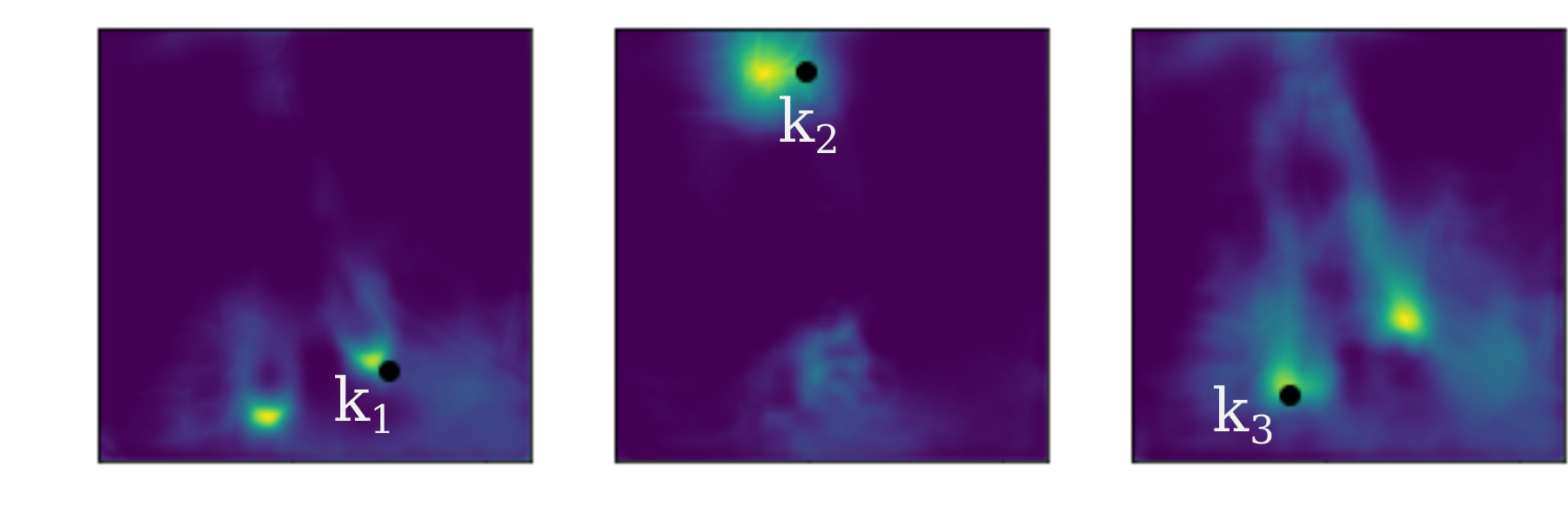}
\includegraphics[width=0.3\columnwidth]{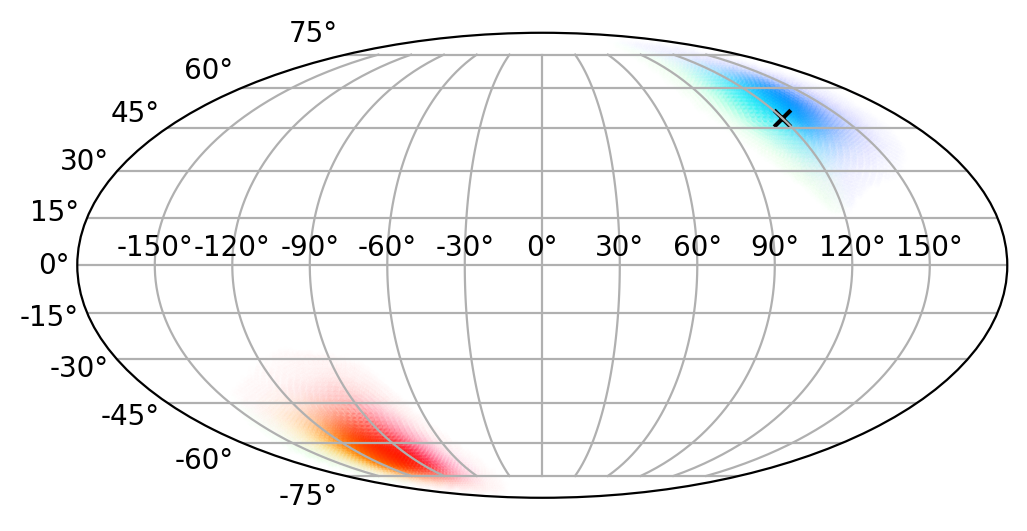}

\includegraphics[width=0.15\columnwidth]{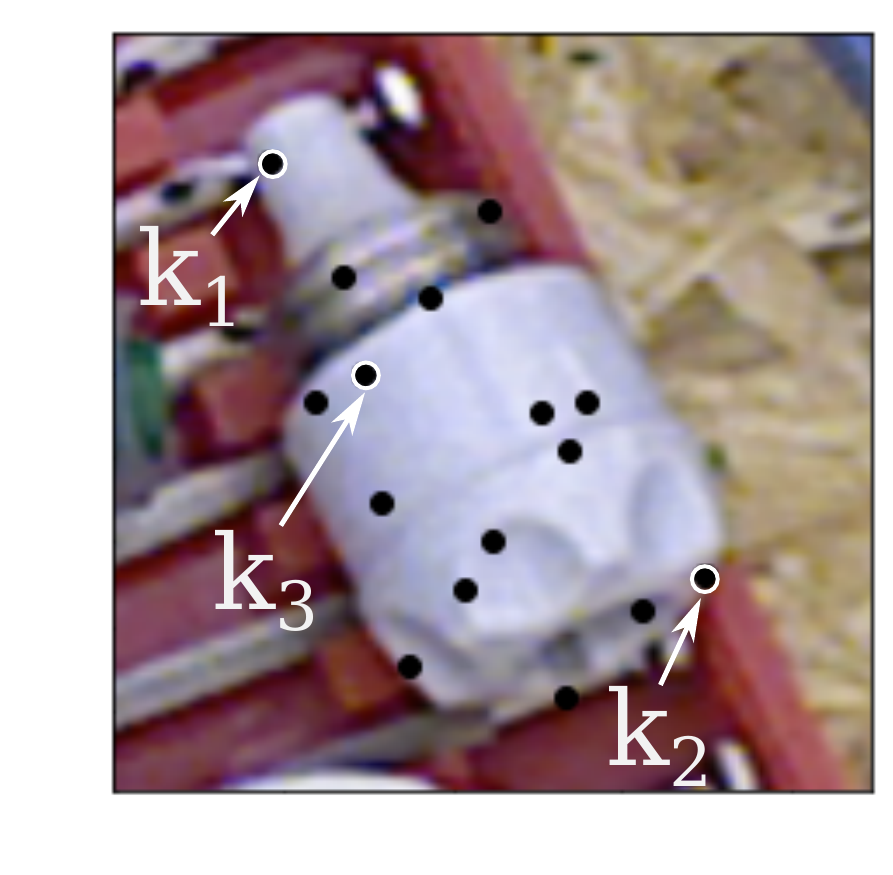}
\includegraphics[width=0.46\columnwidth, trim={0 185 860 0},clip]{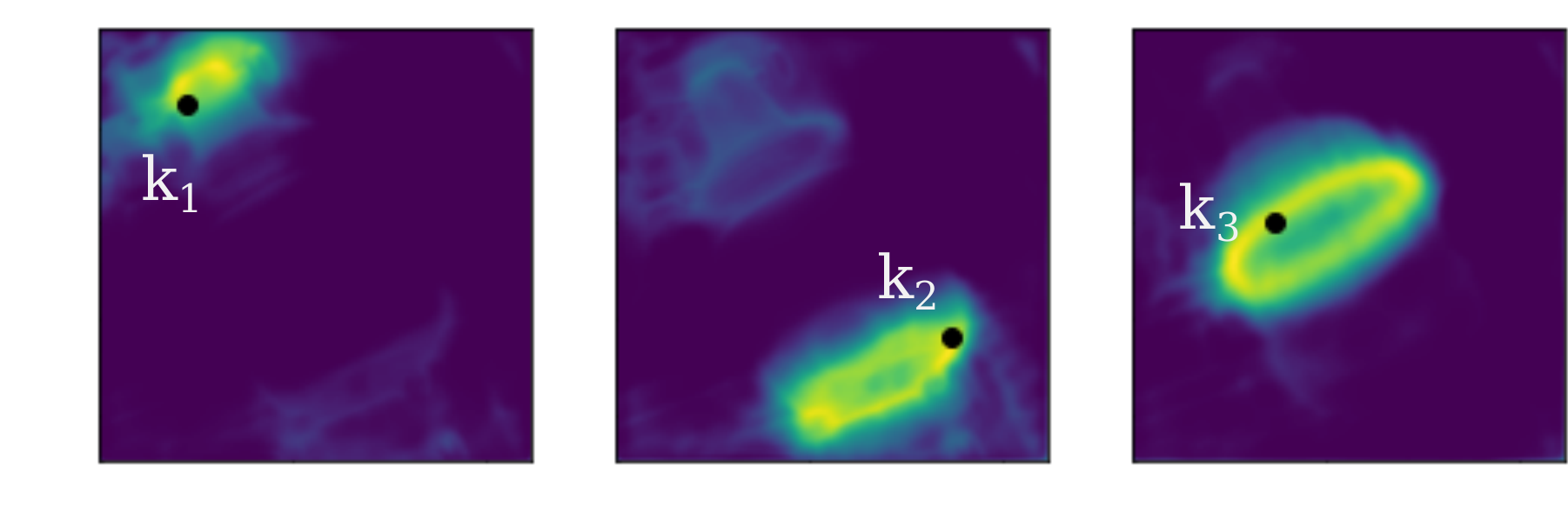}
\includegraphics[width=0.3\columnwidth]{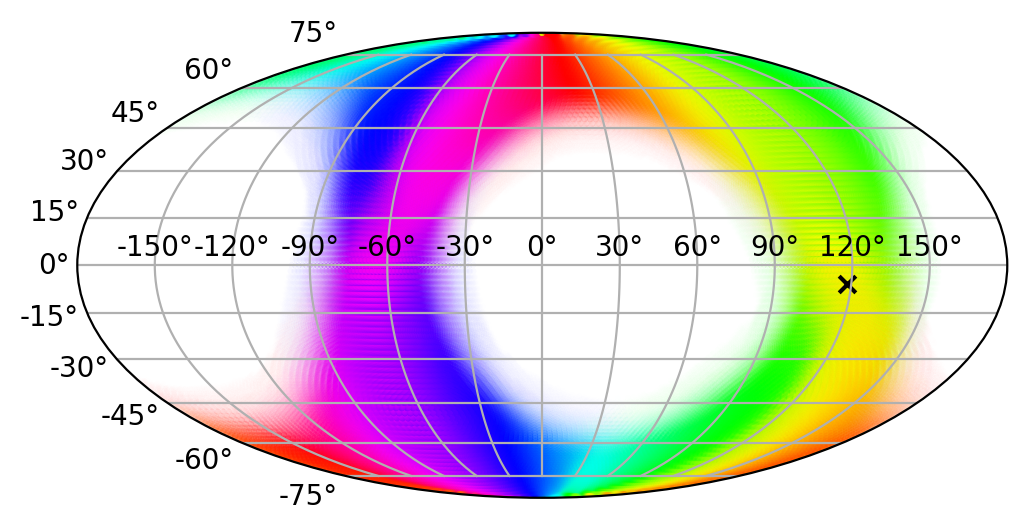}
\caption{Visualization of an image crop, the logarithm of the marginal probability distributions for the first three keypoints, and a visualization of the resulting orientation distribution. The rows shows the YCB-V suger box, YCB-V large clamp, and T-LESS object 1.}
\label{fig:heatmaps}
\end{figure}

Our contribution is a pose distribution estimation method that estimates a probability density function over object pose from a single RGB image. The function is formulated implicitly which enables the predicted distribution to be continuous without being constrained by the descriptive ability of parametric distributions. The formulation is based on an explicit derivation of the probability density function using object keypoints as an intermediary representation of object pose. This increases the interpretability of the estimates as illustrated in Fig. \ref{fig:heatmaps}, which shows estimated keypoint distributions and resulting pose distribution for three different objects. Our method is based on conservative approximations which ensures a high level of reliability.

The scope of the evaluation is limited to rotation distribution estimation. This is done to allow for comparison to existing methods, which primarily focus on estimating the distribution on SO(3). Our method is evaluated on the YCB-V and T-LESS datasets and compared to previously published results of state-of-the-art methods.

The paper is organized as follows: Section \ref{sec:relatedwork} presents related work, followed by section \ref{sec:method} in which our method is presented. The implementation of the method together with the experimental procedure is presented in section \ref{sec:experiments}, and results are shown and discussed in section \ref{sec:results}. Finally section \ref{sec:conclusion} presents the conclusion.

\section{Related work}
\label{sec:relatedwork}
The estimation of object pose from a single RGB image is a topic that has received extensive study by the machine vision community. Our method is inspired by single pose estimation methods that rely on locating object keypoints in images through the estimation of heatmaps, which encode the probability distribution of the keypoints, e.g. \cite{nibali2018numerical}. 

When object keypoints are easily recognizable, such that their uncertainty distribution in the image plane is approximately Gaussian, it is possible to analytically propagate the uncertainty from keypoints to pose~\cite{franaszek2017propagation}. However, the assumption of normality becomes very poor when visual ambiguities are present.

One approach to estimating a pose distribution is by representing the distribution through an ensemble of pose estimates from one or more pose estimation methods~\cite{shi2021fast, okorn2020learning, manhardt2019explaining}. These methods rely on the assumption that the ensemble accurately represents the true pose distribution. However, unless the pose estimators generating the ensemble are explicitly trained for this, it is questionable if the estimated uncertainty distribution reflects the true distribution.

Pose distributions are often multimodal and difficult to model analytically. Therefore, it is common to assume that the rotational and positional part of the distribution is decoupled, assume that the object position is normally distributed, and focus on object rotation estimation. A common approach to rotation distribution estimation is to sample SO(3) and attribute a probability to each unique rotation. This has been done using a confusion matrix constructed from training data \cite{marton2018improving}, treating rotation estimation as a discrete classification problem \cite{su2015render}, or training a network to compute latent features which implicitly encode orientation such that comparison between sampled rotations and a query image is done in latent space. This has both been done with an autoencoder using cosine distance to measure latent feature similarity~\cite{sundermeyer2018implicit, deng2021poserbpf}, or using the latent features from an existing pose estimator combined with an MLP network for regressing unnormalized likelihood~\cite{okorn2020learning}.

An alternative to the sampling approach is to model the pose distribution using parametric distributions. Although it has been argued that the Matrix Fisher distribution is well suited for deep learning-based parameter regression~\cite{mohlin2020probabilistic}, the most common choice of orientation distribution is the Bingham distribution since it explicitly handles the antipodal symmetry of unit quaternions, which are often used to parameterize orientation. Since unimodal distributions are unable to express multiple orientation ambiguities, several works estimate the pose uncertainty as a mixture of Binghams. This has been done by fitting Bingham distributions to the output of a multiple hypotheses pose estimator \cite{manhardt2019explaining}, or direct regression of the Bingham parameters using deep learning \cite{okorn2020learning, deng2022deep, gilitschenski2019deep}. Related is \cite{prokudin2018deep} in which a network is trained to estimate 1D rotation distributions as mixtures of von Mises distributions.

Similar to our method, \cite{murphy2021implicit} proposes to express the probability density function implicitly. In their work, a network is trained end-to-end to output the unnormalized likelihood from an image and a query pose. The likelihood function is then normalized using an equivolumetric grid. Unlike \cite{murphy2021implicit}, our method uses an intermediary representation which increases the interpretability of the pose distribution estimates. Furthermore, our intermediary representation is translation equivariant, which is arguably the most important regularization on the image domain.

\section{Method}
\label{sec:method}
Our method uses object keypoints as a means to estimate object pose. $N$ keypoints are defined in the object's reference frame and a network is trained to estimate the marginal probability distributions over the keypoints' projections on the image plane. These marginal distributions will be referred to as heatmaps. The pose distribution is formulated implicitly, in that given a query pose and the heatmaps, the method returns the unnormalized likelihood of the given pose. The normalized likelihood can then be computed using a sampling-based strategy.

The following three subsections discuss the details of our method. Section \ref{sec:method_keypoints} derives and discusses the relationship between pose distribution and image keypoints as well as the approximations needed when using marginal instead of conditional probabilities. Section \ref{sec:method_network} discusses the training of the network which estimates the heatmaps. Finally, section \ref{sec:method_normalization} discusses the sampling-based method of normalization. 

\subsection{Relationship between pose distribution and keypoints}
\label{sec:method_keypoints}
Pose estimation from a single RGB camera is defined as the estimation of an object's pose, $\theta$, which consists of the rotation, $R\in \text{SO}(3)$, and translation, $t\in\mathbb{R}^3$, relative to the camera from a single RGB image $I$. Pose distribution estimation can then be defined as the estimation of a probability density function over object pose conditioned on the input image, $\text{p}(\theta | I)$.

In the proposed method $N$ object keypoints are used to model the correspondence between an object and its projection on the image plane. The keypoints, $K_i \in \mathbb{R}^3$, are defined relative to the object reference frame. Given an object pose and camera intrinsics the 3D keypoints can be projected onto the image plane resulting in $N$ 2D keypoints, $k_i \in \mathbb{R}^2$. Using the law of total probability, the keypoints as an intermediary representation, and the chain rule, the pose distribution can be reformulated as:

\begin{align}
    \proba{\theta | I} &= \int \proba{\theta,\vek{k}|I}d\vek{k} = \int \proba{\theta | \vek{k},I}\proba{\vek{k}|I}d\vek{k}
    \label{eq:firstIntegral}
\end{align}

where $p(\vek{k}|I)$ is the joint distribution over $k_i$ for $i=1,2...,N$ conditioned on the image, and $p(\theta|\vek{k},I)$ is the distribution over object pose conditioned on both the image and a fixed keypoint configuration. A set of four or more non-colinear keypoints uniquely define an object pose, so the likelihood $p(\theta|\vek{k},I)$ will only be non-zero for the unique pose where the projection of keypoints $\vek{K}$ on the image plane is identical to the keypoint configuration $\vek{k}$. This can be expressed using the Dirac delta function as:

\begin{align}
\proba{\theta | \vek{k},I} = \delta (\veksub{k}{$\uptheta$} - \vek{k})
\label{eq:diracDelta}
\end{align}

where $\veksub{k}{$\uptheta$}$ is the projection of the keypoints $\vek{K}$ using the camera intrinsics and the pose $\theta$. Using Eq. \ref{eq:diracDelta} the integral in Eq.  \ref{eq:firstIntegral} can be rewritten as:

\begin{align}
    \proba{\theta | I} &= \int \delta (\veksub{k}{$\uptheta$} - \vek{k})\proba{\vek{k}|I}d\vek{k} = \proba{\veksub{k}{$\uptheta$}|I}
    \label{eq:poseToKeypoints}
\end{align}




Eq. \ref{eq:poseToKeypoints} states that the likelihood of the object being in pose $\theta$ is equal to the joint likelihood of observing the keypoint configuration $\veksub{k}{$\uptheta$}$.

While estimating the joint distribution over keypoints $\proba{\vek{k}|I}$ is non-trivial, the estimation of the marginal likelihoods, $\proba{k_\text{i}|I}$ are more easily achieved. A common approximation of the joint distribution from marginal distributions is the assumption of independence:

\begin{align}
    \proba{\vek{k}}\approx \proba{k_1}\proba{k_2}...\proba{k_N}
    \label{eq:independent}
\end{align}

However, when the assumption of independence is invalid the approximation can be very poor. The effect is most extreme when all keypoints are strongly coupled such that the value of a single keypoint uniquely defines the rest. If $\veksub{k}{\,-i}$ is used to denote all keypoints excluding $k_i$, the chain rule states that $\proba{\vek{k}} = \proba{\veksub{k}{\,-i}|k_i}\proba{k_i}$. Under strong coupling and assuming a valid configuration $\vek{k}$, this simplifies to $\proba{\vek{k}} = \proba{k_i}$. The product of the marginals in the extreme case thus yields $\proba{k_1}\proba{k_2}...\proba{k_N} = \proba{\vek{k}}^N$. Consequently, under strong coupling, the joint distribution can be computed from the marginals as:

\begin{align}
    p(\vek{k}) = \sqrt[N]{\proba{k_1}\proba{k_2}...\proba{k_N}}
    \label{eq:coupled}
\end{align}

Eq. \ref{eq:independent} and \ref{eq:coupled} express two extremes. The error of using \ref {eq:independent} when the keypoints are strongly coupled is to diminish regions of low probability while using \ref{eq:coupled} when the keypoints are loosely coupled will lead to an attenuation of regions of low probability. As discussed in Sec. \ref{sec:introduction}, it is crucial that estimates are not overconfident, while underconfident estimates are acceptable. For this reason, our method uses the conservative approximation formulated in Eq. \ref{eq:coupled}. Combining Eq. \ref{eq:poseToKeypoints} and \ref{eq:coupled} yields:

\begin{align}
    \proba{\theta | I} \approx
    \hat{\text{p}}(\theta | I) =
        C \sqrt[N]{\prod _{i=1}^N \proba{k_\text{$\theta$,\textit{i}}|I}
    }
    \label{eq:finalApprox}
\end{align}

where $C$ is a normalization constant.

\subsection{Keypoint heatmaps}
\label{sec:method_network}
The proposed method assumes that an object detector is available to provide an $r\times r$ crop of the object, where $r$ is the crop resolution. It is assumed that this crop contains the entire object, both visible and occluded parts, such that all keypoints are inside the crop. The heatmaps are estimated from the crops using a U-Net~\cite{ronneberger2015unet}, with an RGB image as input ($r \times r \times 3$) and the heatmaps as output ($r \times r \times N$), where the i'th channel in the output is an estimate of $\proba{k_i|I}$.

To learn the heatmaps $\proba{k_i | I}$, we represent the problem as spatial classification using cross-entropy loss, where the ground truth class is the nearest pixel to the projected keypoint $k_i$. Identification of the exact nearest pixel may encourage overfitting, so as to avoid this, we regularize the training by instead setting the ground truth heatmap to an isotropic Gaussian centered around the ground truth keypoint, $\proba{k_i | I} = \mathcal{N}(k_i, \sigma)$. The total loss is then a sum over the $N$ cross-entropy losses:

\begin{align}
    L = -\sum_{i=1}^{N} \mathcal{N}(k_i,\sigma)\log \hat{\text{p}}(k_i|I)
\end{align}


\subsection{Normalization}
\label{sec:method_normalization}
The normalization constant $C$ in Eq. \ref{eq:finalApprox} can be approximated by densely sampling SE(3) and choosing $C$ such that $\sum_{\theta_i \in SE(3)}\hat{\text{P}}(\matsub{\theta}{i}|I)=1$, where $\hat{\text{P}}(\matsub{\theta}{i}|I)$ is used to denote the probability mass of sample $i$. However, even when the spatial dimensions are bounded, densely sampling SE(3) requires a very large number of samples, since with $n$ samples along a dimension the computational complexity is $\text{O}(n^6)$. For dense sampling to be feasible, the computation of likelihood must be both fast and parallelizable. Our method has been explicitly developed for this and is capable of computing the likelihood for millions of samples pr. second. This is possible because the image only has to be passed through the network once, after which the computation of likelihood for a single sample only requires the projection of the keypoints followed by a lookup in the heatmaps, all of which are done in parallel on a GPU.

The sampling of SO(3) is done using the method presented in \cite{yershova2010generating} as suggested by \cite{murphy2021implicit}. The method decouples a rotation into the direction of the frame's z-axis expressed as a point on a 2D unit sphere, and a tilt around the z-axis. The unit sphere is sampled in an equiarea grid using the HEALPix method\cite{gorski2005healpix}, while the sampling of the tilt is chosen to make the sides of each volume element equal. This creates an equivolumetric grid with each volume being $V_\text{grid} = \pi ^2 / M$, where M is the number of samples in the grid. The sampling method divides SO(3) recursively and for $s$ recursions the number of samples is $M=72\cdot8^s$. Since the grid is equivolumetric, the likelihoods can be approximated from the probability masses using $\hat{\text{p}}(\matsub{\theta}{i}|I)\approx\hat{\text{P}}(\matsub{\theta}{i}|I)/V_\text{grid}$.
\section{Experiments}
\label{sec:experiments}

\begin{table}[bt]
    \centering
    \begin{tabular}{l|cc|cc|c}
                     &      \multicolumn{4}{c|}{Summary of Evaluation in \cite{okorn2020learning}}                                          &    \\
                     &      I                  &  II             &     III           &      IV             &     Ki-Pode (ours)           \\
\hline
master chef can      &      -0.78                 &  -1.66               &     \good{2.32}    &      0.09                &  \best{  4.36 } \\   
cracker box          &      \good{4.03}           &  3.75                &     -0.09          &      3.71                &  \best{  6.05 } \\   
sugar box            &      3.77                  &  \good{5.94}         &     2.62           &      4.20                &  \best{  6.47 } \\   
tomato soup can      &      0.90                  &  2.02                &     2.52           &      \good{3.99}         &  \best{  5.16 } \\   
mustard bottle       &      3.85                  &  4.61                &     3.02           &      \good{4.81}         &  \best{  5.45 } \\   
tuna fish can        &      \bad{-3.12}           &  -0.20               &     \good{2.64}    &      1.23                &  \best{  5.10 } \\   
pudding box          &      2.18                  &  2.64                &     3.13           &      \good{4.54}         &  \best{  4.91 } \\   
gelatin box          &      4.65                  &  \good{6.25}         &     3.53           &      5.73                &  \best{  6.35 } \\   
        potted meat can      &      1.18          &  \best{\good{3.28}}  &     1.60           &      3.06                &          3.11  \\   
banana               &      2.58                  &  0.58                &     2.27           &      \good{3.70}         &  \best{  4.80 } \\   
pitcher base         &      3.34                  &  4.68                &     2.35           &      \good{4.88}         &  \best{  5.39 } \\   
bleach cleanser      &      3.91                  &  \best{\good{4.70}}  &     2.29           &      3.38                &          4.02   \\   
bowl                 &      \best{\good{1.37}}    &  \bad{-2.77}         &     -1.21          &      \bad{-9.62}         &         -1.50  \\   
mug                  &      3.73                  &  2.50                &     2.75           &      \best{\good{4.72}}  &          4.62  \\   
power drill          &      4.31                  &  \good{5.92}         &     2.43           &      4.17                &  \best{  6.14 } \\   
wood block           &      2.64                  &  -2.09               &     -0.51          &      \best{\good{4.49}}  &         -1.76  \\   
scissors             &      \good{3.74}           &  0.51                &     0.66           &      1.63                &  \best{  5.49 } \\   
large marker         &      \bad{-7.59}           &  -0.29               &     \good{1.02}    &      \bad{-8.13}         &  \best{  3.94 } \\   
large clamp          &      \bad{-5.64}           &  \bad{-6.67}         &     -1.63          &      \bad{-3.54}         &  \best{  3.07 } \\   
extra large clamp    &      \bad{-5.20}           &  \bad{-2.92}         &     \good{0.19}    &      \bad{-5.03}         &  \best{  2.25 } \\   
foam brick           &      -0.26                 &  -2.28               &     \good{1.70}    &      \bad{-12.0}         &  \best{  2.44 } \\   
\hline                                                                                                                                                   
All                  &       0.86                 &   1.71               &     1.74           &      1.43                &  \best{4.09}        \\                  
    \end{tabular}

    \caption{The table shows the mean log-likelihood score for our method as well as the top 4 performing pose distribution methods from \cite{okorn2020learning}. The roman numerals refer to the methods I:~Conf w/DenseFusion, II:~Reg ISO w/DenseFusion, III:~Comp w/PoseCNN, and IV:~Dropout w/PoseCNN. For a description of each method see the original paper. Italic font is used to highlight the best performance among the evaluations in \cite{okorn2020learning}, while bold font highlights the best performance among all methods including ours. Red highlights scores below -2.29 which is worse than a uniform distribution. }
    \label{tab:result_table}
\end{table}

The implementation of the proposed method uses the U-Net implementation from \cite{usuyama2020pytorchunet}, which uses a ResNet-18 backbone pretrained on ImageNet. The number of keypoints is chosen to be $N=16$. The object keypoints $\vek{K}$ are chosen using furthest distance sampling, which ensures that the keypoints are spread out over the surface of the object. The model is trained using physically-based BlenderProc~\cite{denninger2019blenderproc} renderings provided in the YCB-V and T-LESS parts of the BOP-challenge dataset~\cite{hodan2020bop}. The standard deviation of the Gaussian used in the loss function was set to 1px and the training was done for 10 epochs on a GeForce GTX 1080 Ti GPU. To avoid training on images where most of the object is outside the image or the object is occluded, the training is only performed on images for which the object appears in the image and is at maximum 95\% occluded. During training, the images are augmented using random gamma, Gaussian blur, Gaussian noise, ISO noise, and color jitter~\cite{buslaev2020albumentations}, as well as random scaling, rotation, and translation of the image crop. Our implementation of the SO(3) grid sampling method is based on the one provided by~\cite{murphy2021implicit}.

The evaluation focuses on estimating the rotation distribution, $\probahat{R | I}$, such that the proposed method can be compared with existing works. Our method estimates an implicit distribution on SE(3), so the rotational part of the distribution must be marginalized by integrating over the spatial dimensions. The normalization relies on a dense sampling of the pose space so the limiting factor on the sampling density is computation time. In this evaluation, the number of samples has been chosen such that the computation time is well under a second. The resolution of rotation space has been prioritized, so the rotational space has been sampled using 3 HEALPix recursions ($72*8^3=36864$ samples) while the spatial dimensions perpendicular to the camera axis are sampled in an 11x11 grid around an estimated object position (121 samples covering a $10\text{mm} \times 10\text{mm}$ area) for a total of 4.46 million samples. The computation of all samples takes approximately 100 ms on a GeForce GTX 1080 Ti GPU. The marginal rotation distribution is then computed by numerically integrating over grid positions: $\proba{R_i | I} = \int \text{p}({R_i,t|I})dt \approx \left(\sum_{t\in t_\text{grid}} \hat{\text{P}}({R_i,t|I})\right)/V_\text{grid}$

As stated in Sec. \ref{sec:method_network} it is assumed that an object detector provides a bounding box around the object. In the evaluation, the bounding box is chosen to be 20\% larger than a tight crop, and uncertainty on estimation is simulated by using the ground truth bounding box with a random scaling of $\pm5\%$ and a random translation computed such that the object remains fully inside the crop. Furthermore, it is assumed that an estimate of the object position is provided by the object detector, which is simulated by using the ground truth object position with added uniform noise of $\pm10mm$ on all three dimensions. The spatial grid in the normalization is centered around this estimate.

The implementation is evaluated on the test part of the BOP version of the YCB-V and T-LESS datasets. The rotation distribution across the SO(3) grid is estimated and the log-likelihood of the ground truth rotation is chosen as the closest sample in the grid. The mean loglikelihood score (meanLL) for a single object is then computed by averaging across all images of the object and the dataset score is computed by averaging over the meanLL scores for objects in the dataset. The results of the evaluation on the YCBV and T-less datasets are compared to the results published in \cite{okorn2020learning} and \cite{murphy2021implicit} respectively.

The evaluation also presents a qualitative evaluation of our method's interpretability in the form of heatmaps and orientation distributions for three representative objects. The visualization of the rotation distribution is inspired by \cite{murphy2021implicit}. For the visualization, an orientation is parameterized by the direction of the object's canonical x-axis expressed as longitude and latitude $(a,b)$, and the tilt $c$ around the x-axis. The distribution is then visualized by plotting $(a,b)$ using a Mollweide projection, indicating $c$ with color, and indicating the probability mass with the alpha value: $\alpha_i =P_i/P_\text{max}$. The ground truth pose is indicated with an x. The visualization is created using the ground truth crop and object position, and 5 HEALPix recursions.

To investigate the impact of sampling density on the meanLL score, the score has been computed for increasing numbers of recursions of the HEALPix method, using the ground truth crop and object position. The number of samples also determines the theoretical maximum of the log-likelihood, with the limit being when a single sample has a probability mass of one, which corresponds to a log-likelihood of  $V_\text{grid}^{-1}=M/\pi^2$. 
\section{Results}
\label{sec:results}

The evaluation in \cite{okorn2020learning} evaluated 9 different approaches for pose distribution estimation. Each approach was implemented twice using either DenseFusion or PoseCNN as the basis of the implementation, which resulted in a comparison of 18 different methods. Our analysis of the reported results reveals that four of the methods can account for all but one of the per-object best results. The condensed summary shown in Table \ref{tab:result_table} shows that no single method was able to perform well on all objects. The results show that our method is superior for most objects. The relatively low meanLL for the wood block and bowl can be attributed to the many discrete and continuous symmetries respectively. In such cases, there will be many orientations with a low probability that should have been zero. These regions are attenuated due to the approximation of Eq. \ref{eq:coupled}, leading to dispersed probability distributions. It is important to note that unlike method I, II, and IV, our method doesn't have a meanLL score less than -2.29 which correspond to a uniform distribution on SO(3).

The evaluation of different state-of-the-art methods on the T-LESS dataset published in \cite{murphy2021implicit} does not contain per object evaluations, so only the meanLL score over all objects is computed. The scores from \cite{murphy2021implicit} and our method is shown in Table \ref{tab:result_table_tless}. The T-LESS dataset contains many objects with near rotational symmetry, which for our method, as discussed for the bowl in YCB-V, leads to relatively modest performance, presumably because we only model keypoint marginals. Our method is fundamentally different from previous work and can be considered an early work towards interpretable and translational equivariant representations for pose distributions with great potential for further improvements. For instance, an efficient auto-regressive model using this formulation would be able to represent the joint keypoint distributions.

To qualitatively evaluate the interpretability of our method, Fig. \ref{fig:heatmaps} shows image crops, the logarithm of heatmaps, and rotation distribution for three representative objects. The first row shows an object for which most keypoints are well localized, leading to a single well-defined orientation. The second row shows an object with a discrete 180-degree rotation symmetry which in the heatmaps appears as two peaks. The third row shows an object with near rotational symmetry. It is clear from the heatmaps, that the U-Net is able to locate the keypoints as lying on ellipses, but cannot detect that there is only near rotational symmetry. Furthermore, the ellipses are dispersed which leads to uncertainty of the direction of the symmetry axis. The resulting orientation distribution appears as a continuous region on SO(3) as shown in the orientation visualization.

\begin{figure}[bt]
\centering
\hspace{30pt}
\includegraphics[width=0.5\columnwidth]{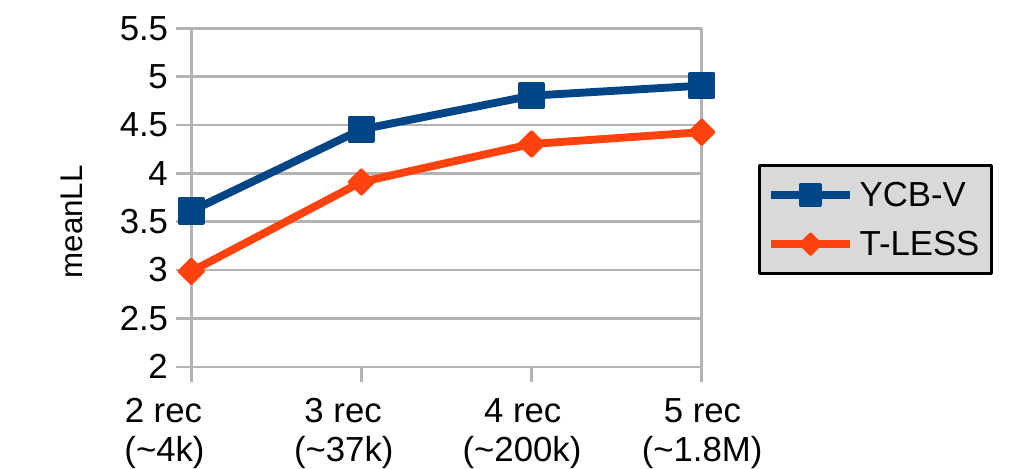}
\caption{The impact of sampling density on the meanLL score. The x-axis shows the number of healpix recursions used to generate the equivolumetric grid on SO(3), and the corresponding number of samples. The theoretical upper bound on the log likelihood for recursions 2-5 is respectively 6.1, 8.2, 10.3, and 12.4.}
\label{fig:loglikelihood_vs_recursions}
\end{figure}

The impact of sample density on the meanLL score is shown in figure \ref{fig:loglikelihood_vs_recursions}. The analysis indicates that increasing sample density improves the score with diminishing returns as the number of samples increases. The computation time scales linearly with the number of samples, which increases by a factor of 8 for each HEALPix recursion. As discussed in Sec. \ref{sec:experiments}, using 3 recursions and an $11\times11$ spatial grid takes approximately 100ms, so a rough estimate of the computation time with 4 or 5 recursions is 0.8s or 6.4s respectively. Ultimately, the choice of how many recursions to use will depend on the available computational budget.

\begin{table}[bt]
    \centering

    \begin{tabular}{lc}
        \hline
        Deng$^1$ & 5.3\\
        Gilitschenski$^1$ & 6.9\\
        Prokudin$^1$  & 8.8\\
        IPDF$^1$ & 9.8\\
        Ki-Pode (ours) & 3.3\\
        \hline
    \end{tabular}

    \caption{MeanLL score on the T-LESS dataset ($^1$Results published in \cite{murphy2021implicit}).}
    \label{tab:result_table_tless}
\end{table}

\section{Conclusion}
\label{sec:conclusion}
We have proposed a novel method for estimating a probability density function over 6D object poses. The distribution is formulated implicitly using keypoints as an intermediary object representation which ensures a high expressiveness of the distribution as well as a high level of interpretability of the estimates. The proposed method approximates the probability distribution from marginal distributions over object keypoints and is based on conservative approximations which leads to high reliability of the results. The method has been evaluated on the YCB-V and T-LESS datasets and has been confirmed to perform reliably on all objects.

\bibliography{references}

\end{document}